\documentclass[10pt,twocolumn,letterpaper]{article}

\usepackage{cvpr}              

\usepackage{graphicx}
\usepackage{amsmath}
\usepackage{amssymb}
\usepackage{booktabs}
\usepackage{multirow}
\usepackage{comment}
\usepackage{caption}
\usepackage{wrapfig}
\usepackage{enumitem}
\usepackage{mathtools}
\usepackage{array}
\usepackage{color}
\usepackage{algorithm}
\usepackage{colortbl}
\definecolor{greyline}{rgb}{0.105,0.410,0.113}
\newcommand{\squishlist}{
	\begin{list}{$\bullet$}
		{ \setlength{\itemsep}{0pt}
			\setlength{\parsep}{1pt}
			\setlength{\topsep}{1pt}
			\setlength{\partopsep}{0pt}
			\setlength{\leftmargin}{1.5em}
			\setlength{\labelwidth}{1em}
			\setlength{\labelsep}{0.5em} } }
\newcommand{\squishend}{\end{list} 
}

\usepackage[toc,page]{appendix}

\usepackage[pagebackref,breaklinks,colorlinks]{hyperref}

\usepackage[capitalize]{cleveref}
\crefname{section}{Sec.}{Secs.}
\Crefname{section}{Section}{Sections}
\Crefname{table}{Table}{Tables}
\crefname{table}{Tab.}{Tabs.}

\def\confName{CVPR}
\def\confYear{2023}

\begin{document}

\title{Learning Procedure-aware Video Representation \\from Instructional Videos and Their Narrations}

\author{Yiwu Zhong$^{1}$\thanks{Work done while Yiwu Zhong was an intern at Meta.}, Licheng Yu$^{2}$, Yang Bai$^{2}$, Shangwen Li$^{2}$, Xueting Yan$^{2}$\thanks{Co-corresponding authors.}, Yin Li$^{1}$\footnotemark[2]\\
$^1$University of Wisconsin-Madison, $^2$Meta AI\\
\texttt{\scriptsize{yzhong52@wisc.edu,\ \{lichengyu, yangbai, dylanwen, xyan18\}@meta.com},\ yin.li@wisc.edu}
}
\maketitle

\begin{abstract}

The abundance of instructional videos and their narrations over the Internet offers an exciting avenue for understanding procedural activities. In this work, we propose to learn video representation that encodes both action steps and their temporal ordering, based on a large-scale dataset of web instructional videos and their narrations, without using human annotations. Our method jointly learns a video representation to encode individual step concepts, and a deep probabilistic model to capture both temporal dependencies and immense individual variations in the step ordering. We empirically demonstrate that learning temporal ordering not only enables new capabilities for procedure reasoning, but also reinforces the recognition of individual steps. Our model significantly advances the state-of-the-art results on step classification (+2.8\%/+3.3\% on COIN / EPIC-Kitchens) and step forecasting (+7.4\% on COIN). Moreover, our model attains promising results in zero-shot inference for step classification and forecasting, as well as in predicting diverse and plausible steps for incomplete procedures. Our code is available at \url{https://github.com/facebookresearch/ProcedureVRL}.
\end{abstract}
\vspace{-0.5em}

\section{Introduction}
\label{sec:intro}

\begin{figure}
    \centering    \includegraphics[width=1.0\linewidth]{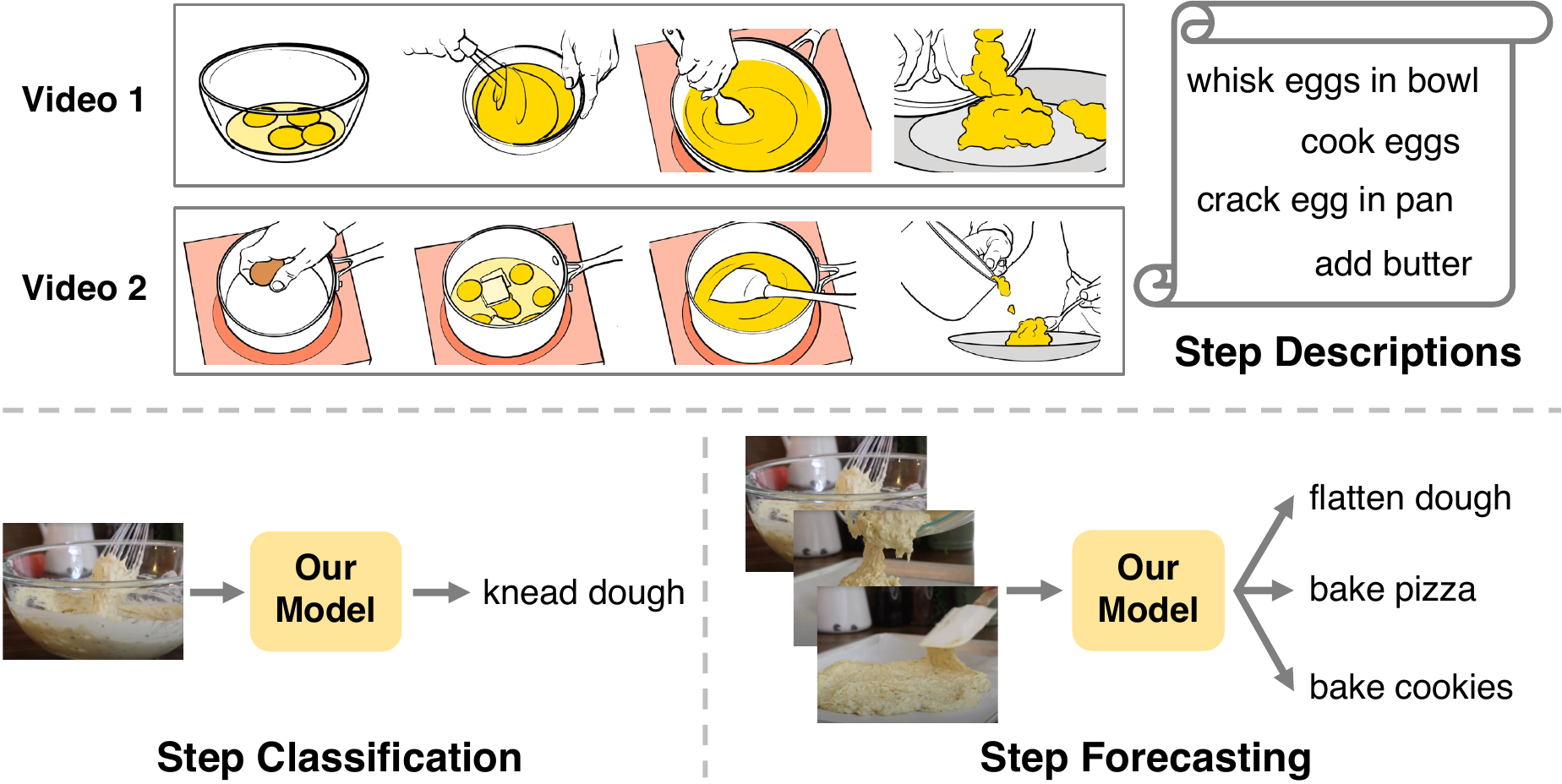} \vspace{-1.5em}
    \caption{\textbf{Top}: During training, our model learns from procedural videos and step descriptions to understand individual steps and capture temporal ordering and variations among steps. \textbf{Bottom}: Once trained, our model supports zero-shot step classification and forecasting, yielding multiple credible predictions.}
    \label{fig:teaser} \vspace{-1.0em}
\end{figure}

Many of our daily activities (\eg cooking or crafting) are highly structured, comprising a set of action steps conducted in a certain ordering. Yet how these activities are performed varies among individuals. Consider the example of making scrambled eggs as shown in Fig.\ \ref{fig:teaser}. While most people tend to whisk eggs in a bowl, melt butter in a pan, and cook eggs under medium heat, expert chefs have recommended to crack eggs into the pan, add butter, and stir them under high heat. Imagine a vision model that can account for the individual variations and reason about the temporal ordering of action steps in a video, so as to infer prior missing steps, recognize the current step, and forecast a future step. Such a model will be immensely useful for a wide range of applications including augmented reality, virtual personal assistant, and human-robot interaction.  

Understanding complex procedural activities has been a long-standing challenge in the vision community~\cite{6126279,5206492,1640961,868686,609450,4624298}. While many prior approaches learn from annotated videos following a fully supervised setting~\cite{Kuehne_2014_CVPR,zhou2018towards,Elhamifar_2019_ICCV}, this paradigm is difficult to scale to a plethora of activities and their variants among individuals. A promising solution is offered by the exciting advances in vision-and-language pre-training, where models learn from visual data (images or videos) and their paired text data (captions or narrations)~\cite{radford2021learning,li2021align,sun2019videobert,zhu2020actbert} in order to recognize a variety of concepts. This idea has recently been explored to analyze instructional videos~\cite{miech2020end,lin2022learning}, yet existing methods are limited to recognize single action steps in procedural activities.

In this paper, we present a first step towards modeling temporal ordering of action steps in procedural activities by learning from instructional videos and their narrations. Our key innovation lies in the joint learning of a video representation aiming to encode individual step concepts, and a deep probabilistic model designed to capture temporal dependencies and variations among steps. The video representation, instantiated as a Transformer network, is learned by matching a video clip to its corresponding narration. The probabilistic model, built on a diffusion process, is tasked to predict the distribution of the video representation for a missing step, given steps in its vicinity. With the help of a pre-trained vision-and-language model~\cite{radford2021learning}, our model is trained using only videos and their narrations from automatic speech recognition (ASR), and thus does not require any manual annotations. 

Once learned, our model celebrates two unique benefits thanks to our model design and training framework. First, our model supports \textit{zero-shot inference} given an input video, including the recognition of single steps and forecasting of future steps, and can be further fine-tuned on downstream tasks. Second, our model allows \textit{sampling multiple video representations} when predicting a missing action step, with each presenting a possibly different hypothesis of the step ordering. Instead of predicting a single representation with the highest probability, sampling from a probabilistic model provides access to additional high-probability solutions that might be beneficial to prediction tasks with high ambiguity or requiring user interactions.

We train our models on a large-scale instructional video dataset collected from YouTube (HowTo100M~\cite{miech2019howto100m}), and evaluate them on two public benchmarks (COIN~\cite{tang2019coin} and EPIC-Kitchens-100~\cite{Damen2021PAMI}) covering a wide range of procedural videos and across the tasks of step classification and step forecasting. Through extensive experiments, we demonstrate that (1) our temporal model is highly effective in forecasting future steps, outperforming state-of-the-art methods by a large margin of \textbf{+7.4\%} in top-1 accuracy on COIN; (2) modeling temporal ordering reinforces video representation learning, leading to improved classification results (\textbf{+2.8\%}/\textbf{+3.3\%} for step classification on COIN/EPIC-Kitchens) when probing the learned representations; (3) our training framework offers strong results for zero-shot step classification and forecasting; and (4) sampling from our probabilistic model yields diverse and plausible predictions of future steps. \smallskip

\noindent \textbf{Contributions}. Our work presents the first model that leverages video-and-language pre-training to capture the temporal ordering of action steps in procedural activities. Our key technical innovation lies in the design of a deep probabilistic model using a diffusion process, in tandem with video-and-language representation learning. 
The result is a model and a training framework that establish new state-of-the-art results on both step classification and forecasting tasks across the major benchmarks.
Besides, our model is capable of generating diverse step predictions and supports zero-shot inference.

\begin{figure*}[h]
	\centering
	\includegraphics[width=0.98\linewidth]{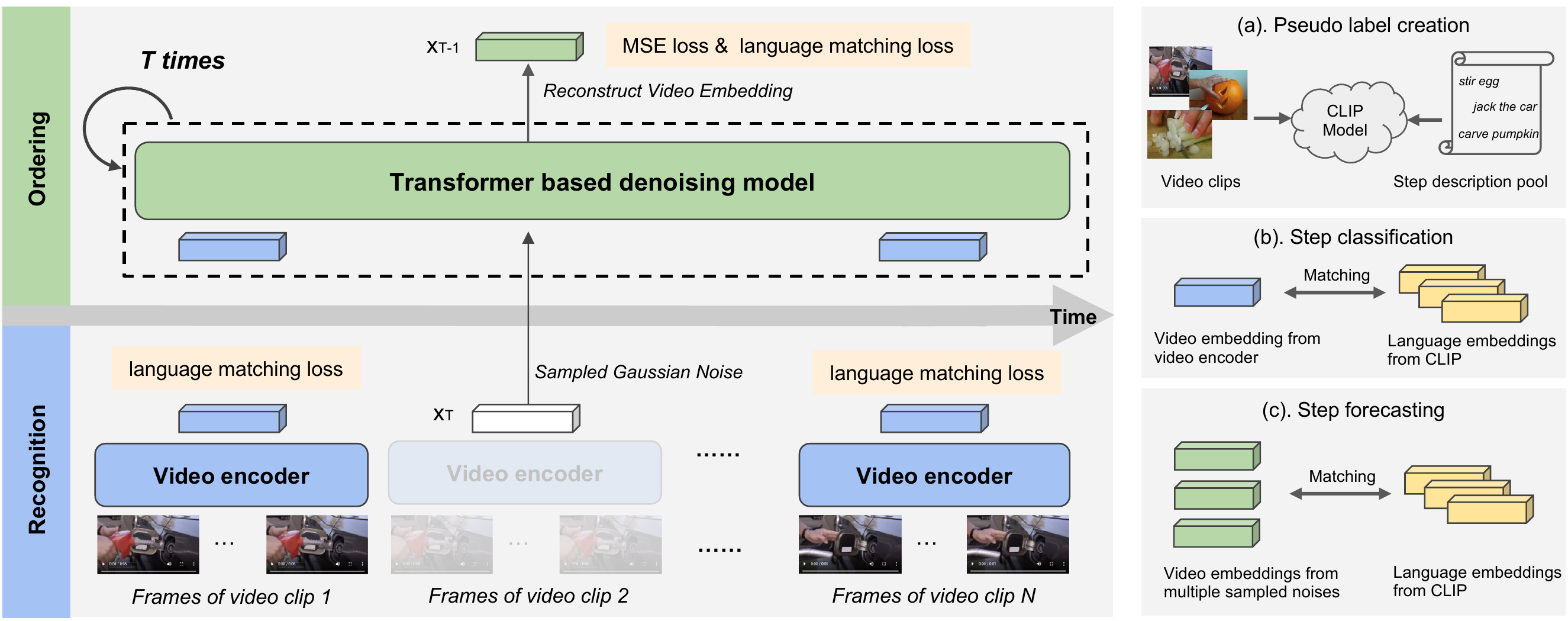} \vspace{-0.5em}
    \caption{Overview of our approach. \textbf{Left panel}: Our model consists of (1) a video encoder that takes a video clip and encodes it into a video embedding; (2) a transformer-based denoising model that samples noises from Gaussian distribution and generates video embedddings conditioned on the embeddings of adjacent video clips. \textbf{Right panel}: We leverage trained image-language model CLIP to create pseudo labels for individual video clips (a). After training, our model supports step classification given an input video clip (b), and step forecasting given a video that records previous steps (c). Note that diverse embeddings can be generated by sampling various noises.}
    \label{fig:model_overview} \vspace{-1.0em}
\end{figure*}

\section{Related Work}
\label{sec:related_work}

\noindent \textbf{Understanding Procedural Activities}. Reasoning about procedural activities, including their action steps and the temporal ordering of these steps, has been a central problem in activity recognition. While early works model temporal ordering with stochastic grammars~\cite{6126279,5206492,1640961,868686,609450,4624298}, more recent works consider supervised learning to localize steps and predict their ordering by learning from videos with human annotated action steps~\cite{Kuehne_2014_CVPR,zhou2018towards,Elhamifar_2019_ICCV,zhukov2019cross,tang2019coin,chang2020procedure,Damen2021PAMI}. To alleviate the burden of costly video annotations, several works propose various forms of weakly supervised settings, with assumptions that the ordered list of steps is given without their temporal boundaries~\cite{bojanowski2014weakly,bojanowski2015weakly,zhukov2019cross,Zhao2022P3IVPP}, or that the key steps and their ordering remain fixed across all videos~\cite{sener2015unsupervised,alayrac2016unsupervised,goel2018learning,kukleva2019unsupervised,Elhamifar_2019_ICCV,Elhamifar20}. 

Most of prior methods focus on the tasks of step classification and localization~\cite{bojanowski2014weakly,bojanowski2015weakly,zhukov2019cross,sener2015unsupervised,alayrac2016unsupervised,kukleva2019unsupervised,Elhamifar_2019_ICCV,Elhamifar20}. Others have considered the tasks of step forecasting~\cite{sener2019zero}, step verification~\cite{qian2022svip} and procedure planning~\cite{Zhao2022P3IVPP}. Our work also seeks to understand procedural activities. Different from these approaches, our method focuses on learning video representation from videos and their narrations without using human annotations. The resulting video representation can be leveraged for step classification and step forecasting. \smallskip

\noindent \textbf{Learning from Procedural Videos and Narrations}. The success of vision-and-language pre-training has fueled a new line of research that seeks to learn concepts of individual steps from instructional videos and their narrations~\cite{malmaud2015s,xu2020benchmark,Shen_2021_CVPR,han2022temporal}. For example, Miech \etal~\cite{miech2020end} propose MIL-NCE to learn representations from instructional videos~\cite{miech2019howto100m} and their narrations extracted using ASR. 

The most relevant work is DistantSup~\cite{lin2022learning}, where they propose using distant supervision from a textual knowledge base (wikiHow)~\cite{koupaee2018wikihow} to denoise text narrations from ASR. Specifically, DistantSup leverages a pre-trained language model~\cite{song2020mpnet} to link step descriptions from wikiHow to text narrations from video ASR results, and thus to create training labels for individual steps in videos. Different from~\cite{lin2022learning}, our method models the temporal ordering of steps in procedural activities, thus moves beyond representations of single steps to support temporal reasoning in videos. Further, our method learns from videos and narrations only, with the help of a pre-trained image-language model~\cite{radford2021learning} yet without using a textual knowledge base.\smallskip

\noindent \textbf{Video-and-Language Pre-Training}. A relevant topic is video-and-language pre-training, aiming at learning video representation from videos and their paired natural language descriptions~\cite{ghadiyaram2019large,xu2021videoclip,bain2021frozen,Lei_2021_CVPR,yang2021taco,sun2019videobert,zhu2020actbert,li-etal-2020-hero,luo2020univl,zellers2021merlot,fu2021violet,wang2022all}, often generated from ASR outputs.  Despite the latest development in ASR, automatically-transcribed speech from videos can be rather noisy and lacks precise temporal alignment with the visual content. Several recent works seek to address this challenge. VideoCLIP~\cite{xu2021videoclip} starts from the pre-trained MIL-NCE model and further improves the model by retrieval augmented training with overlapped video-text pairs. Bain \etal~\cite{bain2021frozen} collect a less noisy dataset of video alt-text pairs and geared the model to match these pairs. Our work shares the key idea of learning from video and text data as prior work, and seeks to leverage external knowledge from a pre-trained image-language model~\cite{radford2021learning}.

Another relevant work is MERLOT~\cite{zellers2021merlot}. While both MERLOT and our work seek to learn video representation, our method differs from MERLOT in two folds. Our method models the sequence order of video clips for understanding procedural activities. MERLOT learns binary relative order between two given video frames for multi-modal reasoning and does not directly support action forecasting. Both methods consider a masked prediction task, yet MERLOT predicts the most likely text embeddings, while our method estimates the distribution of video representations using a deep probabilistic model.\smallskip

\noindent \textbf{Diffusion Models}. Diffusion models~\cite{sohl2015deep,song2020improved} provide a powerful approach to characterize the probability density of high dimensional signals, and have recently demonstrated impressive results on generating high fidelity visual data, such as images~\cite{nichol2021glide,dalle2,saharia2022photorealistic,rombach2022high}, videos~\cite{ho2022video}, and human body motion~\cite{tevet2022human}. Our work adapts diffusion process to model the temporal ordering of steps in procedural videos. In doing so, our method not only facilities the learning of expressive video representations for individual steps, but also enables the anticipation of future action steps.

\section{Method}
\label{sec:method}

We consider the problem of learning video representation for understanding procedural activities from instructional videos and their narrations. An input video is represented as a sequence of $N$ clips $\{v_1, v_2, ..., v_N\}$. Each $v_i$ captures a potential action step in the input video, and the time step $i$ records the temporal ordering of these clips. The video clips $\{v_i\}$ can be either segmented by using the timestamps of ASR outputs (as we consider during training), or densely sampled from a video following their temporal ordering (as we use during inference). During learning, we further assume that an ordered set of sentences $\{s_1, s_2, ..., s_N\}$ is associated with the video clips $\{v_1, v_2, ..., v_N\}$, with each $s_i$ describing the action step in video clip $v_i$. These sentences $\{s_i\}$ can be the output text from ASR, or given by matching the video clips to a text corpus using an external vision language model~\cite{radford2021learning}.\smallskip

\noindent \textbf{Procedural-aware Video Representation}. Our goal is to learn video representation that encodes both action step concepts and their temporal dependencies across a range of procedural activities. Our representation consists of (a) a video encoder $f$ that extracts a representation $x_i$ from an input clip $v_i$ (\ie, $x_i=f(v_i)$); and (b) a probabilistic model that characterizes the conditional probability $p(x_j=f(v_j) | \{x_i=f(v_i)\}_{i\neq j})$ $\forall j$. This design is highly flexible and supports a number of procedural reasoning tasks. $f$ offers video representation suitable to classify individual steps in a clip. $p(x_j | \{x_i\}_{i\neq j})$ models the temporal dependencies among steps, and can be used to predict the video representation of missing steps and further infer their labels.\smallskip 

\noindent \textbf{Method Overview}. To learn our representation, we leverage a pre-trained text encoder $g$ that remains fixed during learning, and extend the idea of masked token modeling, populated in natural language processing~\cite{kenton2019bert}. For each input video and its narrations at training time, we randomly sample a clip $v_j$ from $\{v_1, v_2, ..., v_N\}$ and mask it out. We then train our model to predict the distribution of $x_j=f(v_j)$ from $\{x_i=f(v_i)\}_{i\neq j}$ (\ie, $p(x_j | \{x_i\}_{i\neq j})$), align the expectation of the predicted distribution $\mathbb{E}(x_j) $ with the corresponding text embedding $y_j=g(s_j)$, and match all other video representations $\{x_i=f(v_i)\}_{i\neq j}$ to their text embeddings $\{y_i=g(s_i)\}_{i\neq j}$. 

Despite the conceptual similarity, our learning is fundamentally different from masked token prediction. Our method seeks to characterize the distribution of $x_j$ instead of predicting the most likely $x_j$, resulting in a more principled approach to capture the temporal dependencies among steps, as well as the new capability of sampling multiple high-probability solutions for $x_j$. Our method is illustrated in Fig.\ \ref{fig:model_overview}. In what follows, we lay out the formulation of our model, and describe its training and inference schemes. 

\subsection{Modeling Action Steps and Their Ordering}

Formally, given an input video with its clips $\{v_1, v_2, ..., v_N\}$ and their narrations $\{s_1, s_2, ..., s_N\}$, our method assumes a factorization of $p(Y=\{y_i\} | X=\{x_i\})$ with video representation $x_i=f(v_i)$ (learnable) and text embedding $y_i=f(s_i)$ (pre-trained and fixed).
\begin{equation}
\small
\resizebox{.43\textwidth}{!}{
$p(Y | X) = p(y_j | x_j) \cdot p(x_j | \{x_i\}_{i \neq j}) \cdot \prod_{i} p(y_i | x_i) \quad \forall j.$
}\label{eq:model}
\end{equation}
$p(y_i | x_i)$ measures the alignment between a video representation $x_i$ and a text embedding $y_i$. $p(x_j | \{x_i\}_{i \neq j})$ characterizes the distribution of a video representation for a missing step given the representations of all other steps, thereby modeling the temporal ordering of steps. Note that our model is not limited to single step prediction and can be readily extended to predict multiple missing steps.\smallskip

\noindent\textbf{Matching Image and Text Representations}. Our model matches the video representation $x_i$ and text embedding $y_i$ in a learned vector space, such that the alignment between them can be measured by cosine similarity. We will later instantiate this definition into a more tractable form for learning. Yet it suffices to notice that $p(y_j | x_j)$ does not involve additional learnable parameters given $x_i$ and $y_i$.\smallskip

\noindent\textbf{Modeling Step Ordering with Diffusion Process}. The key challenge lies in the modeling of $p(x_j | \{x_i\}_{i \neq j})$, as the video representation $x_i$ is at least of a few hundred dimensions. To this end, we propose to model $p(x_j | \{x_i\}_{i \neq j})$ using a diffusion process~\cite{sohl2015deep,song2020improved} conditioned on observed video representations $\{x_i\}_{i \neq j}$. Here we briefly describe diffusion process in the context of our model, and refer the readers to recent surveys for more technical details~\cite{croitoru2022diffusion,yang2022diffusion}.  

Specifically, we assume a diffusion process that gradually adds noise to the input $x_j$ over $t\in [0, 1, ..., T]$ steps. 
\begin{equation}
\small
\begin{split}
p(x^{1:T}_j | x^0) &= \prod_{t=1}^T p(x^t_j|x^{t-1}_j), \\
p(x^t_j|x^{t-1}_j) &= \mathcal{N}(\sqrt{\alpha_{t}}x^{t-1}_j, (1-\alpha_t)I).
\end{split}
\label{eq:diffusion}
\end{equation}
where $\alpha_t$ are constant hyper-parameters. The reverse diffusion (denoising) process is parameterized with
\begin{equation}\resizebox{.41\textwidth}{!}{
$p(x^{0:T}_j | \{x_i\}_{i \neq j}) = p(x_j^T | \{x_i\}_{i \neq j}) \prod_{t=1}^T p_\theta(x^{t-1}_j | x^t_j,  \{x_i\}_{i \neq j})$.
}
\end{equation}
With sufficiently large $T$, $p(x_j^T | \{x_i\}_{i \neq j}) \sim \mathcal{N}(0, I)$, \ie a standard Gaussian noise that is independent of $\{x_i\}_{i \neq j}$. The denoising process is approximated by $p_\theta(x^{t-1}_j | x^t_j,  \{x_i\}_{i \neq j})$ using a neural network with parameters $\theta$ such that
\begin{equation}\resizebox{.41\textwidth}{!}{
$p_\theta(x^{t-1}_j | x^t_j,  \{x_i\}_{i \neq j}) = \mathcal{N}(\mu_\theta(x^t_j,  \{x_i\}_{i \neq j}), \Sigma_\theta(x^t_j,  \{x_i\}_{i \neq j}))$
}
\end{equation}
In practice, we follow Ho \etal~\cite{ho2020denoising} and Tevet \etal~\cite{tevet2022human} to directly predict $x^0_j$ by using a denoising model $h$. With slight abuse of the symbols, we denote 
\begin{equation}
\small
\hat{x}_j^0 = h(x_j^t, \{x_i\}_{i \neq j}, t).
\end{equation}
$h$ is realized using a Transformer network with the embedding of step $t$ as part of its inputs. Once learned, one can sample from $\mathcal{N}(0, I)$ and apply $h$ through the denoising process to predict $x_j$ based on $\{x_i\}_{i \neq j}$.

\subsection{Learning from Videos and Their Narrations}

Our training approximately maximizes the likelihood of Eq.\ \ref{eq:model} given a set of training videos and their narrations.\smallskip

\noindent \textbf{Pseudo Labels from CLIP}. It is straightforward to directly align video representations to the embeddings of their corresponding ASR text. Doing so, however, faces the challenges of low-quality ASR text and imprecise alignment between video and ASR sentences.  
To address these challenges, we propose to create pseudo labels by leveraging a pre-trained image-language model (\eg, CLIP~\cite{radford2021learning}).

Specifically, we first create a pool of step descriptions in the form of verb phrases (\eg, ``add water'', ``wear gloves'') parsed from ASR sentences~\cite{Shen_2021_CVPR}, with their embeddings as $\{y^{1:K}\}$. Then a trained CLIP model is applied to link each video clip with verb phrases, by matching the averaged visual features across frames with the language embeddings of verb phrases. The resulting matching scores are used as our training target. 

Our pseudo labeling instantiates the matching $p(y_i | x_i)$ between video representation and text embedding using
\begin{equation}
\small
p(y_i | x_i) = \mathrm{softmax}\left(\frac {x_i^{T}  y_i } {\tau \|x_i\|  |y_i\| }\right),
\label{eq:similarity}
\end{equation}
where $y_i$ is selected from pool of verb phrases, \ie $y_i \in \{y^{1:K}\}$, and $\tau$ is the pre-defined temperature. The matching problem thus is converted into a ``classification'' problem, making the training feasible.\smallskip

\noindent \textbf{Learning Objective and Training Loss}.
Our training minimizes an evidence upper bound of the negative log likelihood $- \log p(Y | X)$ in Eq.~\ref{eq:model}. The detailed derivation of evidence upper bound is described in the Appendix. Our objective function constitutes three loss terms:
\begin{equation}
\small
\mathrm{L} = \mathrm{L}_{\mathrm XE} + \mathrm{L}_{\mathrm MSE} + \mathrm{L}_{\mathrm MC}.
\end{equation}

The \textit{first} term $\mathrm{L}_{\mathrm XE}$ seeks to match observed video representation $\{x_i\}$ to their text embeddings $\{y_i\}$, given by 
\begin{equation}
\small
\resizebox{.28\textwidth}{!}{
$\mathrm{L}_{\mathrm XE} = \frac{1}{N-1} \sum_i H\left(p'_i,  p(y_i | x_i)\right),$
}
\label{loss:XE}
\end{equation}
where $H(\cdot, \cdot)$ is the cross entropy, and $p'_i$ are soft targets given by CLIP matching scores. $p(y_i | x_i)$ defined in Eq.~\ref{eq:similarity} measures the similarity between $x_i$ and $y_i$.

The \textit{second} term $\mathrm{L}_{\mathrm MSE}$ comes from the Kullback–Leibler (KL) divergence within our diffusion model, and is computed as
\begin{equation}
\small
\mathrm{L}_{\mathrm MSE} = \mathbb{E}_{x_j^0{\sim}p\left(x_j^0 | {\{x_i\}_{i \neq j}}\right),\ t{\sim}[1,T]} \left[ \|{x_j^0} - \hat{x}_j^0 \|_2^2  \right].
\label{loss:MSE}
\end{equation}
Note that unlike a standard diffusion model, our model directly predicts $\hat{x}_j^0$. This term is applied at each step $t$.

The \textit{third} term $\mathrm{L}_{\mathrm MC}$ is derived from matching the predicted video representation $\hat{x}_j^0$ to its text embedding $y_j$.
\begin{equation}
\small
\mathrm{L}_{\mathrm MC} =  \mathbb{E}_{x_j^0{\sim}p\left(x_j^0 | {\{x_i\}_{i \neq j}}\right),\ t{\sim}[1,T]} \left[ H\left(p'_j, p(y_j | \hat{x}_j^0)\right) \right], 
\label{loss:MC}
\end{equation}
where $p'_j$ are again soft targets given by CLIP model, and $\hat{x}_j^0$ is denoised from a sampled noise. During training, we adopt Monte Carlo estimation for $\mathbb{E}_{p} \left[ - \log p(y_j|\hat{x}_j^0) \right]$, by minimizing $- \log p(y_j|\hat{x}_j^0)$ for each sampled $\hat{x}_j^0$. We attach this term at each step $t$.

A critical design choice lies in $p(y_j | \hat{x}_j^0)$. $p(y_j | \hat{x}_j^0)$ is simplified into a score function between a video representation and a finite set of text embeddings (defined using verb phrases). This allows us to reach our loss terms without worrying about global normalization constant as commonly encountered in energy-based models. Indeed, $H(p'_j, p(y_j | \hat{x}_j^0))$ can be interpreted as providing guidance by matching video to text embeddings. This term thus resembles the key idea of classifier guidance, which has shown to be helpful for learning diffusion models~\cite{dhariwal2021diffusion}.

\subsection{Model Inference}

Once trained, our model offers a procedure-aware representation with two key components. First, the video encoder $f(\cdot)$ serves as a feature extractor for any input video clips. Second, the diffusion model, represented as its denoising model $h(\cdot)$, captures the temporal dependencies among steps. Our representation naturally supports a number of tasks. Here we demonstrate how our model can be used for step classification and step forecasting.\smallskip 

\noindent \textbf{Step Classification}. An input video clip $v$ can be encoded using $f(\cdot)$. The video representation $x=f(v)$ can be directly compared to the text embeddings~\cite{radford2021learning}, so as to support zero-shot step classification. Alternative, an additional classifier can be attached on top of $f$ and further fine-tuned to recognize the action step in input clip. \smallskip  

\noindent \textbf{Step Forecasting}. A future video clip feature $x_{j}$ can be sampled from the diffusion model by drawing from a Gaussian distribution and denoising using $h(\cdot)$. The predicted $x_{j}$ can be further classified into action steps. This prediction can be done using again Monte Carlo estimation given by 
\begin{equation}
\mathbb{E}_{x_j{\sim}p(x_j | {\{x_i\}_{i \neq j}})} \left[ p(y_j | x_j) \right].
\label{expectation_inference}
\end{equation}
Specifically, a noise $x_T$ is first sampled from Gaussian distribution and our denoising model gradually denoises it. At each step $t$, the denoising model $h$ takes a noisy $x_j^t$, predicts clip feature $\hat{x}_j^0$, and diffuses it to $x_j^{t-1}$ based on the sampled noise $x_j^T$, as demonstrated in Eq.~\ref{eq:diffusion}. After $T$ iterations, the predicted clip feature at $t=0$ is used to match text embeddings. By sampling noises for multiple times, we can estimate the most likely $y_j$.

However, sampling can be costly. In practice, to obtain top-1 prediction for missing steps, we adopt approximate inference, where the sampled noise is replaced with a fixed zero vector, corresponding to peak in the Gaussian distribution. Our empirical results validate that approximate inference achieves a very close performance as the expectation over multiple sampled noises.

\section{Experiments and Results}
\label{sec:experiments}

\begin{table*}[]
\centering
\resizebox{0.75\textwidth}{!}{%
\begin{tabular}{l|l|l|l|ll}
\toprule
&\multirow{2}{*}{Model} & \multicolumn{2}{c|}{Pretraining} & \multicolumn{2}{c}{Top-1 Acc. (\%)}   \\
& & \multicolumn{1}{l|}{Supervision} & Dataset & Zero-shot & Fine-tuning\\ \midrule
1 & SlowFast~\cite{Feichtenhofer_2019_ICCV} & \multicolumn{1}{l|}{Supervised: action labels} & Kinetics &  --  & 25.6 \\
2 & TimeSformer~\cite{gberta_2021_ICML} & \multicolumn{1}{l|}{Supervised: action labels} & Kinetics & -- & 34.7 \\
3 & S3D~\cite{xie2018rethinking} & \multicolumn{1}{l|}{Unsupervised: ASR w. MIL-NCE~\cite{miech2020end}} & HT100M & -- & 28.1 \\
4 & TimeSformer~\cite{gberta_2021_ICML} & \multicolumn{1}{l|}{Unsupervised: ASR w. MIL-NCE~\cite{miech2020end}} & HT100M & -- & 34.0 \\
5 & DistantSup~\cite{lin2022learning} & \multicolumn{1}{l|}{Unsupervised: ASR + wikiHow} & HT100M & -- & 39.4 \\ \midrule
6 & Random Guess & -- & \multicolumn{1}{l|}{--} & 0.1 
& -- \\
7 & CLIP~\cite{radford2021learning} & \multicolumn{1}{l|}{Unsupervised: captions} & CLIP400M & 9.4 
& -- \\
8 & Ours & \multicolumn{1}{l|}{Unsupervised: ASR} & HT100M &  \textbf{11.3}  
& \textbf{46.8} \\ \midrule
9& Ours (oracle-5) & \multicolumn{1}{l|}{Unsupervised: ASR} & HT100M &  14.7 & 51.8 \\ \bottomrule 
\end{tabular}
}
\vspace{-0.5em}
\caption{Step forecasting on COIN dataset. We compare to a set of strong baselines and a oracle protocol built on our method.}
\label{tab:step_forecasting}
\vspace{-1em}
\end{table*}

In this section, we first introduce datasets, evaluation protocols and implementation details. Then we demonstrate our results on step forecasting and step classification benchmarks. Finally, we show our qualitative results and conduct ablations to study our model components. 
\smallskip

\noindent \textbf{Datasets.} For {\it pre-training}, we consider HowTo100M dataset~\cite{miech2019howto100m} with 130K hours of YouTube tutorial videos. 
The videos cover various daily tasks, such as foods, housework, vehicles, \etc. 
We use a language parser~\cite{Shen_2021_CVPR} to extract the verb phrases from ASR sentences of these videos and keep 9,871 most frequent verb phrases. 
For {\it fine-tuning}, we train our model on COIN dataset~\cite{tang2019coin,coin_pami} and EPIC-Kitchens-100 dataset~\cite{Damen2021PAMI}, respectively. 
COIN has 476 hours of YouTube videos covering 180 tasks, such as dishes, vehicles, housework, \etc. Human annotators summarize 778 unique steps in total (\eg, ``stir the egg''), and annotate the temporal boundary and the category of each step in all videos. 
EPIC-Kitchens-100 dataset~\cite{Damen2021PAMI} has 100 hours of egocentric videos, capturing daily activities in kitchen. 
Each action in the videos is annotated with an action label and a noun label.
There are 97/300 unique actions/nouns in total.
We use the human annotations in COIN and EPIC-Kitchens-100 to evaluate the model performance.\smallskip

\noindent \textbf{Evaluation Protocols}. Our evaluation considers zero-shot and fine-tuning settings for step classification and step forecasting on COIN and EPIC-Kitchens-100 datasets.
Zero-shot setting indicates that no human annotation is used during pre-training. The pre-trained model is directly tested on the evaluation dataset.
Fine-tuning setting further fine-tunes the pre-trained model using human annotations of action steps. 
For a fair comparison, we follow the same fine-tuning schemes as previous work in respective benchmarks.\smallskip

\noindent \textbf{Implementation Details}. We adopted TimeSformer~\cite{gberta_2021_ICML} as our video encoder, and used the Transformer~\cite{vaswani2017attention} from CLIP's text encoder as denoising model. We set the maximum step $T$ to 4, maximum length of video sequence as 9, and the number of Transformer layers as 4. We used a trained CLIP model (ViT-B/16) to create pseudo labels and encode step descriptions. Following DistantSup~\cite{lin2022learning}, for pre-training we used SGD for 5 epochs and then AdamW~\cite{loshchilov2017decoupled} for 25 epochs with 128 videos in a batch. For fine-tuning, we used AdamW for 15 epochs with batch size of 64. Temperature $\tau$ was set to 0.02. Additional implementation details can be found in Appendix.

\subsection{Step Forecasting}

\begin{table*}[]
\centering
\resizebox{0.79\textwidth}{!}{%
\begin{tabular}{l|l|ll|ll}
\toprule
& \multirow{2}{*}{Model} & \multicolumn{2}{c|}{Pretraining} & \multicolumn{2}{c}{Top-1 Acc. (\%)} \\
& & \multicolumn{1}{l|}{Supervision} & Dataset & \multicolumn{1}{l}{Zero-shot} & Fine-tuning \\ \midrule
1 & TSN (RGB+Flow)~\cite{tang2019coin} & \multicolumn{1}{l|}{Supervised: action labels} & Kinetics & \multicolumn{1}{l}{--} & 36.5* \\
2 & S3D~\cite{xie2018rethinking} & \multicolumn{1}{l|}{Unsupervised: ASR w. MIL-NCE~\cite{miech2020end}} & HT100M & \multicolumn{1}{l}{--} & 37.5* \\ \midrule
3 & SlowFast~\cite{Feichtenhofer_2019_ICCV} & \multicolumn{1}{l|}{Supervised: action labels} & Kinetics & \multicolumn{1}{l}{--} & 32.9 \\
4 & TimeSformer~\cite{gberta_2021_ICML} & \multicolumn{1}{l|}{Supervised: action labels} & Kinetics & \multicolumn{1}{l}{--} & 48.3 \\
5 & ClipBERT~\cite{Lei_2021_CVPR} & \multicolumn{1}{l|}{Supervised: captions} & COCO+VG & \multicolumn{1}{l}{--} & 30.8 \\
6 & VideoCLIP~\cite{xu2021videoclip} & \multicolumn{1}{l|}{Unsupervised: ASR} & HT100M & \multicolumn{1}{l}{--} & 39.4 \\
7 & TimeSformer~\cite{gberta_2021_ICML} & \multicolumn{1}{l|}{Unsupervised: ASR w. MIL-NCE~\cite{miech2020end}} & HT100M & \multicolumn{1}{l}{--} & 46.5 \\
8 & DistantSup~\cite{lin2022learning} & \multicolumn{1}{l|}{Unsupervised: ASR + wikiHow} & HT100M & \multicolumn{1}{l}{--} & 54.1 \\
9 & DistantSup$\dagger$~\cite{lin2022learning} & \multicolumn{1}{l|}{Unsupervised: ASR + wikiHow} & HT100M & \multicolumn{1}{l}{10.2} & 46.6 \\
10 & CLIP~\cite{radford2021learning} & \multicolumn{1}{l|}{Unsupervised: captions} & CLIP400M & \multicolumn{1}{l}{14.8} & 45.9 \\
11 & Ours & \multicolumn{1}{l|}{Unsupervised: ASR} & HT100M & \multicolumn{1}{l}{\textbf{16.6}} & \textbf{56.9} \\ \bottomrule
\end{tabular}
}
\vspace{-0.5em}
\caption{Step classification on COIN dataset. DistantSup$\dagger$ is re-implemented based on their official code base. It is a variant reported in their paper that pre-trains the model to match language embeddings. * indicates the model is fully fine-tuned on COIN dataset.}
\vspace{-0.5em}
\label{tab:step_classification}
\end{table*}

\begin{table*}[]
\centering
\resizebox{0.85\textwidth}{!}{%
\begin{tabular}{l|l|l|l|ccc}
\toprule
&Model & Pretraining Supervision & Pretraining Dataset & Action (\%) & Verb (\%) & Noun (\%)\\ \midrule
1&TSN~\cite{wang2016temporal} & -- & -- & 33.2 & 60.2 & 46.0 \\ 
2&TRN~\cite{zhou2018temporal} & -- & -- & 35.3 & 65.9 & 45.4 \\ 
3&TBN~\cite{kazakos2019epic} & -- & -- & 36.7 & 66.0 & 47.2 \\ 
4&MoViNet~\cite{kondratyuk2021movinets} & -- & -- & \textbf{47.7} & \textbf{72.2} & 57.3 \\ 
5&TSM~\cite{lin2019tsm} & Supervised: action labels & Kinetics & 38.3 & 67.9 & 49.0 \\ 
6&SlowFast~\cite{Feichtenhofer_2019_ICCV} & Supervised: action labels & Kinetics & 38.5 & 65.6 & 50.0 \\ 
7&ViViT-L~\cite{arnab2021vivit} & Supervised: action labels & Kinetics & 44.0 & 66.4 & 56.8 \\ 
8&TimeSformer~\cite{gberta_2021_ICML} & Supervised: action labels & Kinetics & 42.3 & 66.6 & 54.4 \\ 
9&DistantSup~\cite{lin2022learning} & Unsupervised: ASR + wikiHow & HT100M & 44.4 & 67.1 & 58.1 \\ 
10&Ours & Unsupervised: ASR & HT100M & \textbf{47.7} & 69.5 & \textbf{60.3} \\ \bottomrule
\end{tabular}
}\ 
\vspace{-0.5em}
\caption{Step classification on EPIC-Kitchens-100 dataset with fine-tuning setting. Our method outperforms the close competitors (TimeSformer, DistantSup), with results on par with even stronger backbone models (MoViNet).}
\label{tab:epic_step_classification} \vspace{-1em}
\end{table*}

\noindent \textbf{Setup}. We follow the benchmark in DistantSup~\cite{lin2022learning} to evaluate step forecasting on COIN, where top-1 accuracy is reported.
Given a video with previous steps, the model anticipates the category of next single step (\eg, “stir the egg”). This task thus requires explicit modeling of the temporal ordering among steps.
We only fine-tune the diffusion model while keeping the video encoder frozen. 
\smallskip

\noindent \textbf{Results}. Table~\ref{tab:step_forecasting} compares results of our method with a series of baselines. The closest competitor is DistantSup~\cite{lin2022learning} in L5, which learns from ASR text and an external textual knowledge base~\cite{koupaee2018wikihow} using the same video backbone (TimeSformer~\cite{gberta_2021_ICML}). We also include other baselines reported in DistantSup, \eg, SlowFast~\cite{Feichtenhofer_2019_ICCV},  TimeSformer~\cite{gberta_2021_ICML}, and S3D~\cite{xie2018rethinking} from L1 to L4, where models are supervised using human-annotated action labels or video ASR text. Our model significantly outperforms all baselines by at least \textbf{7.4\%} for the fine-tuning setting (\eg, 46.8\% in L8 vs. 39.4\% in L5). Further, we consider a strong baseline for the zero-shot setting by re-purposing CLIP model~\cite{radford2021learning} to match the input video with the descriptions of all step candidates.
Comparing L7 and L8, our model outperforms this variant of CLIP by a clear margin (11.3\% vs. 9.4\%).

A unique property of our model is its ability to output multiple, potentially different predictions. We further evaluate the upper bound of our results by assuming an oracle ranking function that always selects the correction prediction from 5 outputs sampled from our model (Ours (oracle-5)). This oracle further improves the top-1 accuracy from 11.3\% to 14.7\% in L9, suggesting that our model is able to produce diverse predictions for step forecasting.

\subsection{Step Classification}
\noindent \textbf{Setup}. Besides step ordering, we also evaluate step classification on COIN and EPIC-Kitchens-100 datasets, where a model is tasked to classify a trimmed video clip into one of the step categories. 
For COIN, we follow DistantSup~\cite{lin2022learning} to only fine-tune the additional linear layer on top of the pre-trained video encoder. For EPIC-Kitchens-100, we fully fine-tune the video encoder, following~\cite{kondratyuk2021movinets,arnab2021vivit,lin2022learning}. We report the accuracy of step classification on COIN, and that of verb, noun, and action on EPIC-Kitchens-100.\smallskip

\noindent \textbf{Results}.  Table.~\ref{tab:step_classification} summarizes the results on COIN. 
We consider baselines as in DistantSup from L1 to L8 (\eg, SlowFast~\cite{Feichtenhofer_2019_ICCV}, VideoCLIP~\cite{xu2021videoclip}), where models are trained using either action labels or video ASR text.  
To support zero-shot inference, we re-implement a model variant (DistantSup$\dagger$) described in~\cite{lin2022learning}. 
This model is pre-trained to match video embeddings with language embeddings and thus supports recognizing arbitrary step descriptions in L9.
We also report the results of CLIP~\cite{radford2021learning}, which creates the pseudo labels for our pre-training in L10. 
As shown, our model consistently outperforms all the other methods by a clear margin under different settings.
For example, ours outperforms CLIP by \textbf{1.8\%} in zero-shot setting (16.6\% in L11 vs.\ 14.8\% in L10), and outperforms DistantSup by \textbf{2.8\%} (56.9\% in L11 vs.\ 54.1\% in L8) in fine-tuning setting.

Table.~\ref{tab:epic_step_classification} presents our results on EPIC-Kitchens-100.
While TimeSformer~\cite{gberta_2021_ICML} in L8 and DistantSup~\cite{lin2022learning} in L9 use the same video encoder architecture as ours, our model in L10 achieves a clear gain over them, \eg, \textbf{+3.3\%/2.2\%} for action/noun.
The only exception is the lower accuracy (-2.7\%) on verb when compared with MoViNet (MoViNet-A6) in L4, a heavily optimized video backbone.

\begin{figure*}[t]
	\centering
	\includegraphics[width=0.95\linewidth]{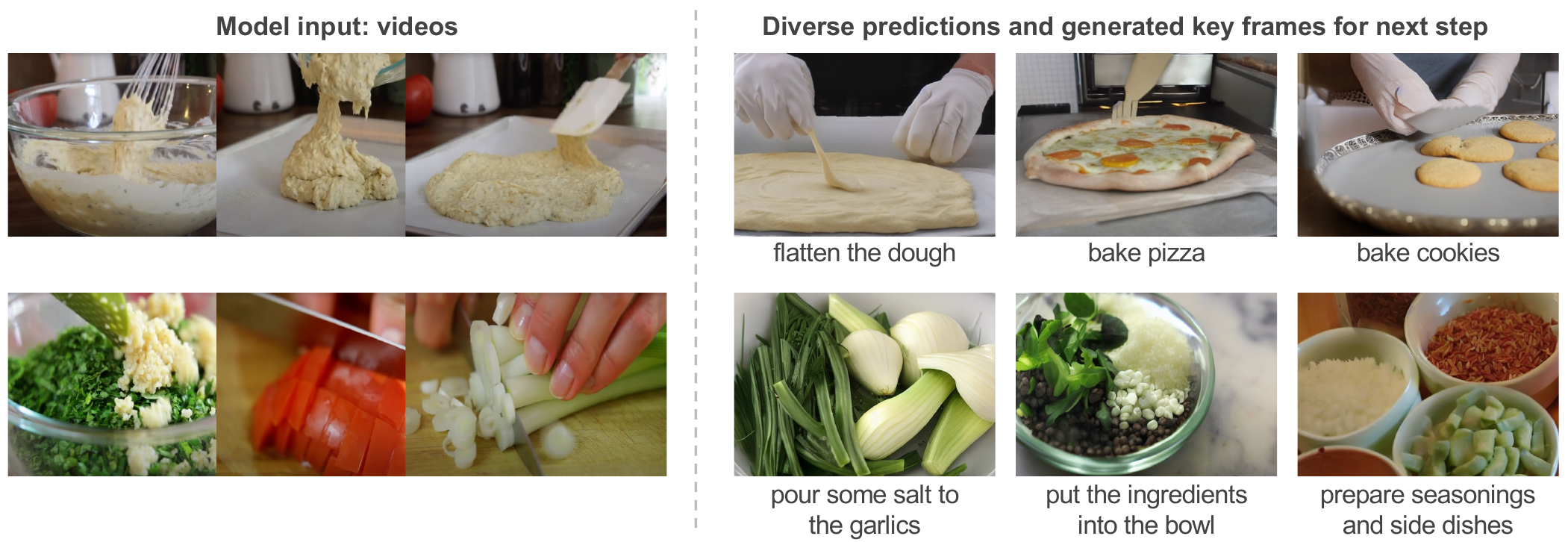} \vspace{-0.5em}
    \caption{Visualization of \textbf{zero-shot} step forecasting and \textbf{key frame generation}. Without using any human annotation during training, our trained model is directly evaluated on COIN dataset~\cite{coin_pami}. Given a video recording previous steps (left), our model is capable of forecasting multiple reasonable predictions and each predicted step is further used for key frame generation (right). We adopt stable diffusion~\cite{rombach2022high} for key frame generation, taking inputs as a text description of step and a sampled frame from input video.}
    \label{fig:visualization}  \vspace{-0.5em}
\end{figure*}

\begin{table*}[]
\centering
\resizebox{0.92\textwidth}{!}{%
\begin{tabular}{c|c|c|cc|cc}
\toprule
\multirow{2}{*}{Model} & \multirow{2}{*}{Pre-training task} & \multirow{2}{*}{Order modeling} & \multicolumn{2}{c|}{Zero-shot (top-1 acc. \%)}              & \multicolumn{2}{c}{Fine-tuning (top-1 acc. \%)}        \\
  &  &  & Step classification  & Step forecasting  & Step classification & Step forecasting \\ \midrule
Ours & Matching & -- & 13.7 & No zero-shot ability & 52.8 & 41.6  \\
Ours & Matching + Ordering & Mask & 16.9  &  10.6  &  56.4  &  43.2  \\ 
Ours   & Matching + Ordering & Diffusion & 16.6 & 11.3  & 56.9  &  46.8  \\ 
\bottomrule
\end{tabular}
}
\vspace{-0.5em}
\caption{Ablation study of pre-training tasks and order modeling. Our proposed order pre-training not only enables zero-shot forecasting, but also significantly improves zero-shot step classification and the fine-tuning results across evaluation tasks. Our diffusion models further improves mask modeling, especially on fine-tuning step forecasting. 
}
\vspace{-1em}
\label{tab:ablation}
\end{table*}

\subsection{Predicting Diverse Future Steps} 
One of the defining characteristics of our model is that it allows us to sample multiple predictions of video representation corresponding to a future step. This leads to an interesting question about the diversity of the predictions, as partially evaluated in our prior experiments. Here we present further demonstration of this capability by visualizing the step forecasting results, and more interestingly, using these results to generate future video frames.  

Fig.\ \ref{fig:visualization} presents the visualization for zero-shot step forecasting and key frame generation. In this setting, our model is pre-trained without any human annotation and is directly tested for step forecasting. 
We show multiple predictions sampled from our diffusion model. Further, we demonstrate that the text description of predicted step can be used to generate the key frames by leveraging the stable diffusion model~\cite{rombach2022high}. 
To keep the generated images visually consistent with the input video, we let stable diffusion model take one input video frame and the description of predicted step as input and generate an image.

As shown in Fig.\ \ref{fig:visualization}, our model is capable of forecasting multiple, reasonable next steps (\eg, ``flatten the dough'', ``bake pizza''), based on which credible future frames can be generated. These results suggest that our model not only predicts meaningful video representations of individual steps, but also captures the variations in step ordering

\subsection{Ablation Studies}

We conduct ablation study on COIN, including step classification/forecasting with zero-shot/fine-tuning setting. Additional ablation results can be found in Appendix.\smallskip

\noindent \textbf{Does modeling of temporal order help?} In Table~\ref{tab:ablation}, we conduct a comparison on two different pre-training tasks: (1) pre-training by only matching video representations to text embeddings of the verb phrases; and (2) pre-training by our method that combines matching and temporal order modeling. In comparison to pre-training using matching only, our method significantly improves the performance for both zero-shot and fine-tuning settings and across step classification and step forecasting tasks. For example, zero-shot step classification is improved from \textbf{13.7\%} to \textbf{16.6\%}. Our results after fine-tuning attains a major gain of +3.6\% and +1.8\% for step classification and step forecasting, respectively. Importantly, our method also enables zero-shot step forecasting by predicting future video representations. These results suggest that our procedure-aware pre-training can effectively facilitate the learning for both video representation and step ordering in procedure activities.\smallskip

\noindent \textbf{Masked Prediction vs.\ Diffusion Model}. We explore a model variant using the mask prediction, sharing similar spirit as the well-known Masked Language Modeling in BERT~\cite{kenton2019bert}. 
Specifically, this variant is trained to recover the video embeddings of masked video clips so that it can match to the assigned verb phrases. Our diffusion model largely outperforms the results of this variant, especially on the step forecasting (+0.7\% and +3.6\% for zero-shot and fine-tuning, respectively, as in Table~\ref{tab:ablation}). 
This result indicates that our diffusion model is a more suitable way to capture the variation inside the step ordering.\smallskip

\noindent \textbf{Approximate Inference}. In Table~\ref{tab:approximation}, we validate that our approximate inference with a single zero-vector can achieve close empirical results as the Monte Carlo estimation (\eg, within 0.1\% difference). Monte Carlo estimation computes the weighted average of multiple sampled predictions (\eg, from 5 sampled noises). 
We run the experiment for 5 times and the variation is small (\eg, $\pm$0.03\%). 
Further, if we assume an oracle ranking function to pick the correct one from sampled predictions, our results can be further boosted by 3.4\% on average, suggesting diverse predictions from our model and ample room to improve.

\begin{table}[]
\centering
\resizebox{0.32\textwidth}{!}{%
\begin{tabular}{l|l|l}
\toprule
\multirow{2}{*}{Model} & \multirow{2}{*}{Inference Type} & \multirow{2}{*}{Top-1 Acc. (\%)} \\
 &   &  \\ \midrule 
Ours & Approximation & 11.33 \\  
Ours & Expectation & 11.34 $\pm$ 0.03 \\  
Ours & Oracle & 14.73 $\pm$ 0.13 \\
\bottomrule
\end{tabular}
}
 \vspace{-0.5em}
\caption{Ablation study of inference schemes. ``Approximation'' uses a zero-vector as noise, ``Expectation'' adopts Monte Carlo estimation, and ``Oracle'' further assumes an oracle ranking function to pick the correct prediction derived from multiple noises.} \vspace{-1.5em}
\label{tab:approximation}
\end{table}

\section{Conclusion}

In this work, we presented a model and a training frame work for learning procedure-aware video representation from a large-scale dataset of instructional videos and their narrations, without the need for human annotations. The key strength of our model lies in the joint learning of a video encoder capturing concepts of action steps, as well as a diffusion model reasoning about the temporal dependencies among steps. We demonstrated that our model achieves strong results on step classification and forecasting in both zero-shot and fine-tuning settings and across COIN and EPIC-Kitchens-100 datasets. We believe our work provides a solid step towards understanding procedural activities. We hope that our work will shed light on the broader problem of video-language pre-training.

{\small
\bibliographystyle{ieee_fullname}
\bibliography{egbib}
}

\clearpage

\setcounter{figure}{0}
\setcounter{table}{0}
\setcounter{section}{0}
\renewcommand{\theequation}{\Alph{equation}}
\renewcommand{\thefigure}{\Alph{figure}}
\renewcommand{\thesection}{\Alph{section}}
\renewcommand{\thetable}{\Alph{table}}

\appendix
\addappheadtotoc
\begin{appendices}

In appendices, we describe (1) the derivation of our training loss, (2) the implementation details of data pre-processing, our model architecture, and key frame generation, (3) experiment results on the additional benchmark on COIN dataset, and (4) additional ablation studies on open-vocabulary recognition, the effects of using ASR phrases and backbone architecture. For sections, figures, tables, and equations, we use numbers (\eg, Sec.\ 1) to refer to the main paper and capital letters (\eg, Sec.\ A) to refer to this appendix.

\begin{table*}[h]
\begin{equation}
\resizebox{0.70\textwidth}{!}{$
\begin{aligned}
- \log p(Y | X) & = - \log (p( y_j|x_j) \cdot p(x_j|\{x_i\}_{i \neq j}) \cdot \prod_{i} p(y_i | x_i))\\
&  \leq  \underbrace{\sum_{i} - \log (p(y_i | x_i))}_{ \text{cross-entropy loss ($\mathrm{L}_{\mathrm XE}$) } }  \\
& +  \underbrace{\sum_{t=1}^{T} \mathbb{E}_{ x_j^t \sim p( x_j^t | x_j^0, \{x_i\}_{i \neq j} ) } \left[ \mathbb{D}_{KL} ( p(x_j^{t-1} | x_j^t, x_j^0, \{x_i\}_{i \neq j}) || p_{\theta}(x_j^{t-1} | x_j^t,  \{x_i\}_{i \neq j}) ) \right]}_{ \text{diffusion model loss ($\mathrm{L}_{\mathrm MSE}$})} \\
& +  \underbrace{\mathbb{E}_{x_j \sim p_{\theta}(x_j|\{x_i\}_{i \neq j})} \left[ - \log ( p( y_j|x_j) )  \right]}_{ \text{cross-entropy loss ($\mathrm{L}_{\mathrm MC}$}) }   
\end{aligned}
$} 
\label{likelihood}
\end{equation}
\end{table*}

\section{Derivation of Training Loss}
Our method aims at minimizing the negative log likelihood $- \log p(Y | X)$ (Eq.~\ref{eq:model} in paper). Here, we provide the derivation of its evidence lower bound, as shown in Eq.~\ref{likelihood}, 
where $x_i$ are video embeddings learned by our video encoder $f(\cdot)$, $y_i$ are text embeddings offered by a pre-trained text encoder $g(\cdot)$ from CLIP~\cite{radford2021learning} that remains fixed during our training. $\{x_i\}$ and $\{y_i\}$ are observed video and text embeddings, while $x_j$ and $y_j$ are the missing (masked) video and text embeddings. 

There are three terms in the evidence lower bound, with each one corresponding to a loss in our main paper. First, $p(y_i | x_i)$ is computed by Eq.~\ref{eq:similarity} of the paper, as a softmax over the cosine similarity between an input video embedding and a set of text embeddings. This term corresponds to the loss $\mathrm{L}_{\mathrm XE}$ (Eq.~\ref{loss:XE}). Second, $p(x_j | \{x_i\}_{i \neq j})$ is approximated using a diffusion model that consists of a diffusion process and an reverse diffusion (denoising) process. This term is performed by the loss $\mathrm{L}_{\mathrm MSE}$ (Eq.~\ref{loss:MSE}). Third, $p(y_j | x_j)$ seeks to predict text embedding $y_j$ using the masked video embedding $x_j$. It is again calculated by Eq.~\ref{eq:similarity} of the paper. This term corresponds to $\mathrm{L}_{\mathrm MC}$ (Eq.~\ref{loss:MC}).

\section{Additional Implementation Details}
\noindent \textbf{Data Pre-processing}: During pre-training, we used the timestamps of ASR sentences to segment video clips from full videos. For step classification, the video clips are trimmed by human-annotated step boundaries. When evaluating step classification, multi-view augmentation is applied with 3 clips sampled on the temporal dimension. For step forecasting (both training and evaluation), we cropped 68 seconds of video before the target action and uniformly cut it into 8 video clips as the model input. For HowTo100M~\cite{miech2019howto100m} and COIN dataset~\cite{tang2019coin,coin_pami}, we sampled 1 frame per second. For EPIC-Kitchens-100 dataset~\cite{Damen2021PAMI}, we sampled 16 frames per second. The text embedding of each verb phrase was the averaged embedding over  28 action prompts\footnote{https://github.com/openai/CLIP/blob/main/data/prompts.md\#kinetics700}.

\noindent \textbf{Model Architecture and Hyper-parameters}: We adopted TimeSformer architecture~\cite{gberta_2021_ICML} for our video encoder. TimeSformer is a Transformer~\cite{vaswani2017attention} based model that applies attention mechanism over both spatial and temporal dimension. For denoising model, we used Transformer from CLIP's implementation\footnote{https://github.com/openai/CLIP} with bi-directional attention. In denoising model, we implemented the maximum time level $T$ as 4, maximum length of video sequence as 9, and the number of Transformer layers as 4. For time variable in diffusion model, we first mapped it into vector representation using position embeddings and then added it to the input of Transformer. When calculating the matching score between video and text embedding (Eq.\ 4 in main paper), we divided the matching score by a temperature $\tau=0.02$ when computing the softmax.

\noindent \textbf{Details about Future Key Frame Generation}: Future key frame generation is posed as text guided image-to-image translation, where the text is provided by our predicted step and the image is from a sampled frame within the current video. Specifically, we use a pre-trained stable diffusion model\footnote{https://github.com/CompVis/stable-diffusion} and employ SDEdit~\cite{meng2021sdedit}. SDEdit adds noise to the sampled input video frame, and then denoises the resulting image using stable diffusion model and the text of our predicted step, in order to generate a future video frame.

\section{Additional Benchmarks}

\subsection{Procedural Activity Classification}
We follow the benchmark in DistantSup~\cite{lin2022learning} to evaluate procedural activity recognition on COIN with top-1 accuracy reported.
Given a video that has recorded multiple steps, the model classifies the entire video into an activity category (e.g., “make coffee”). 
Similar to step forecasting, we only fine-tune the diffusion model to predict activity category, with the frozen video encoder as a feature extractor.

In Table~\ref{tab:procedure_classification}, we compare our model with a series of baselines as in DistantSup~\cite{lin2022learning}, such as SlowFast~\cite{Feichtenhofer_2019_ICCV}, TimeSformer~\cite{gberta_2021_ICML} and S3D~\cite{xie2018rethinking}. These baselines are pre-trained by either human-annotated action labels or video ASR sentences. Our closest competitor is DistantSup~\cite{lin2022learning} which learns individual action concepts by leveraging an external text knowledge base (wikiHow). Our model clearly outperforms all baseline models by a large margin (\eg, \textbf{+1.9} over DistantSup in L8). Our experimental results suggest that our order pre-training approach, which captures the order among steps, can also improve the recognition of the entire sequence of steps, even if it was not designed for this task.

\begin{table*}[]
\centering
\resizebox{0.70\textwidth}{!}{%
\begin{tabular}{l|l|ll|l}
\toprule
& \multirow{2}{*}{Model} & \multicolumn{2}{c|}{Pretraining} & \multicolumn{1}{c}{Top-1} \\
& & \multicolumn{1}{l|}{Supervision} & Dataset & Acc. (\%)  \\ \midrule
1 & TSN (RGB+Flow)~\cite{tang2019coin} & \multicolumn{1}{l|}{Supervised: action labels} & Kinetics &  73.4* \\
2 & S3D~\cite{xie2018rethinking} & \multicolumn{1}{l|}{Unsupervised: ASR w. MIL-NCE~\cite{miech2020end}} & HT100M & 70.2* \\ \midrule
3 & SlowFast~\cite{Feichtenhofer_2019_ICCV} & \multicolumn{1}{l|}{Supervised: action labels} & Kinetics & 71.6 \\
4 & TimeSformer~\cite{gberta_2021_ICML} & \multicolumn{1}{l|}{Supervised: action labels} & Kinetics & 83.5 \\
5 & ClipBERT~\cite{Lei_2021_CVPR} & \multicolumn{1}{l|}{Supervised: captions} & COCO+VG & 65.4 \\
6 & VideoCLIP~\cite{xu2021videoclip} & \multicolumn{1}{l|}{Unsupervised: ASR} & HT100M & 72.5 \\
7 & TimeSformer~\cite{gberta_2021_ICML} & \multicolumn{1}{l|}{Unsupervised: ASR w. MIL-NCE~\cite{miech2020end}} & HT100M & 85.3 \\
8 & DistantSup~\cite{lin2022learning} & \multicolumn{1}{l|}{Unsupervised: ASR + wikiHow} & HT100M & 88.9 \\
9 & Ours & \multicolumn{1}{l|}{Unsupervised: ASR} & HT100M & \textbf{90.8} \\ \bottomrule
\end{tabular}
}
\vspace{-0.5em}
\caption{Procedural activity classification on COIN dataset. * indicates the model is fully fine-tuned on COIN dataset.}
\vspace{-0.5em}
\label{tab:procedure_classification}
\end{table*}

\section{Additional Ablation Studies}

We present additional ablation studies on our model. The experiment settings follow the ablation study in the main paper, unless otherwise noticed.
\smallskip

\begin{figure}[h!]
	\centering
	\includegraphics[width=0.99\linewidth]{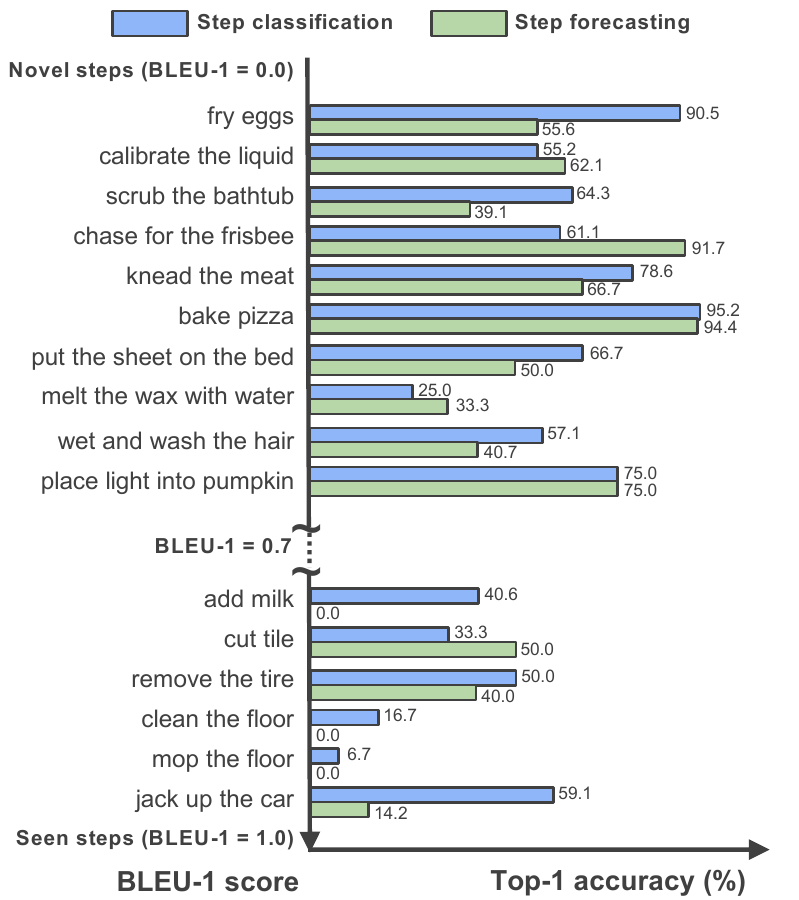} \vspace{-0.5em}
    \caption{Per-category top-1 accuracy for zero-shot step classification and forecasting. We rank the step concepts in COIN dataset by calculating its maximum BLEU-1 score~\cite{papineni2002bleu} versus all step descriptions used in pre-training. }
    \label{openset_chart}
\end{figure}

\begin{table}
\centering
\resizebox{0.49\textwidth}{!}{%
\begin{tabular}{l|p{6cm}}
\toprule
COIN steps & Step descriptions during pre-training \\
\midrule
fry eggs & fry chicken, lay eggs \\ \midrule
calibrate the liquid  &  calibrate the meter \\ \midrule
scrub the bathtub  & clean the bathroom \\ \midrule
chase for the frisbee  &  -- \\ \midrule
knead the meat & cut the meat, cook the meat \\ \midrule
bake pizza  &  bake soda, bake powder, make pizza \\ \midrule
put the sheet on the bed  &  take a sheet, make a sheet  \\ \midrule
melt the wax with water  &  melt the plastic, melt the cheese, put wax \\ \midrule
wet and wash the hair  &  moisturize hair, rinse hair, wet my brush \\ \midrule
place light into pumpkin  &  place your lights, adjust the light \\
\bottomrule
\end{tabular}
} \vspace{-0.5em}
\caption{Visualization of step concepts. We show COIN steps (left) and the step descriptions in pre-training (HowTo100M) that have common verb/noun (right).} \vspace{-0.5em}
\label{openset_samples}
\end{table}

\noindent \textbf{Can our model identify open-vocabulary step concepts?} Part of our learning objective is to match the video representations with text embeddings. Such a design allows our model to support zero-shot recognition as we demonstrated in the paper. One natural question is how well our model performs during zero-shot recognition when facing step concepts that have not been seen during pre-training.

Figure~\ref{openset_chart} measures the overlap between step concepts during pre-training (from ASR results on HowTo100M) and during zero-shot recognition (from human-annotated categories on COIN), and reports per-category results for both seen and novel step categories. Specifically, we adopt BLEU-1 score~\cite{papineni2002bleu} to match the step concepts, and report per-category top-1 accuracy for zero-shot step classification and forecasting. BLEU-1 score as zero indicates the novel steps and BLEU-1 score as one suggests that the exact steps have been seen during pre-training. In addition, we show the steps that have a common verb/noun as COIN steps in Table~\ref{openset_samples}.

We find that our model achieves high accuracy even if facing novel steps, \ie the steps have low BLEU-1 score (\eg, 90.5\% for ``fry eggs''). Further, we compute the top-1 accuracy for the steps with high BLEU-1 scores (\eg, $\ge 0.7$) and the steps with low BLEU-1 scores (\eg, $< 0.7$). These two groups include 103 and 675 steps, respectively, and have close top-1 accuracy across tasks (\eg, 15.9 vs.\ 16.7 for step classification, 14.2 vs.\ 10.9 for step forecasting). These results suggest that our model is not limited to the step concepts considered in pre-training and supports open-vocabulary step recognition. We conjecture that our model has learned the components from similar phrases (\eg, ``fry chicken'' and ``lay eggs'' shown in Table~\ref{openset_samples}), by learning to project video embeddings into the semantic space defined by the text embeddings of CLIP. \smallskip

\begin{table}[]
\centering
\resizebox{0.48\textwidth}{!}{%
\begin{tabular}{c|cc|cc}
\toprule
\multirow{2}{*}{Source}  & \multicolumn{2}{c|}{Zero-shot}              & \multicolumn{2}{c}{Fine-tuning}        \\
   & Classification  & Forecasting  & Classification &  Forecasting \\ \midrule
wikiHow sentences & 11.6 & 8.3 & 48.6 & 38.0 \\
ASR phrases & 11.8 & 9.0 & 47.8 & 38.9 \\ 
\bottomrule
\end{tabular}
} \vspace{-0.5em}
\caption{Ablation study on different sources of step descriptions. Top-1 accuracy (\%) on COIN dataset is reported. All models are pre-trained on a subset of HowTo100M dataset, defined by~\cite{gberta_2021_ICML,lin2022learning}. 
} \vspace{-0.5em}
\label{tab:step_source}
\end{table}

\noindent \textbf{Are ASR phrases sufficient to learn step concepts?}
We propose to use the step phrases parsed from video ASR sentences for learning step concepts. The latest work DistantSup~\cite{lin2022learning} found that external text corpus for procedure activities (\eg, wikiHow~\cite{koupaee2018wikihow}) can largely reduce the noise in ASR sentences. In this section, we explore using wikiHow sentences to pre-train our model.

In Table~\ref{tab:step_source}, we compare our model with a variant pre-trained using wikiHow sentences, following~\cite{lin2022learning}. Our results demonstrate that ASR phrases are sufficient to achieve competitive results across tasks and settings (\eg, +0.7/+0.9 for step forecasting across zero-shot and fine-tuning settings). In other word, our model only requires ASR phrases generated from audio transcriptions of videos, without the need of an external text corpus describing the procedural activities as in~\cite{lin2022learning}.
\smallskip

\noindent \textbf{Backbone Architecture of Video Encoder}. In Table~\ref{tab:backbone_arch}, we study the effects of backbone architectures for our video encoder. We replace the default backbone TimeSformer with MViT-S~\cite{Li_2022_CVPR} which is also a widely-used architecture for video encoders. 
We slightly increase the frame sampling rate of MViT-S from the default value of 4 to 6 so that the encoder can take a longer video (e.g., on COIN, the average duration of a step is 14 seconds).
TimeSformer consistently outperforms MViT-S across tasks (\eg, +4.1 on step classification). 
We conjecture that TimeSformer, which samples 8 frames from consecutive 256 frames, is better suited for recognizing actions with long durations, such as COIN steps. Conversely, MViT-S, which samples 16 frames from consecutive 96 frames, may perform better for recognizing actions with short durations and high-speed motion.

\begin{table}[]
\centering
\resizebox{0.37\textwidth}{!}{%
\begin{tabular}{c|cc}
\toprule
\multirow{2}{*}{Source}  & \multicolumn{2}{c}{Zero-shot}   \\
   & Classification  & Forecasting \\ \midrule
Ours (TimeSformer) & 16.6 & 11.3 \\
Ours (MViT-S) & 12.5 & 9.0  \\ 
\bottomrule
\end{tabular}
} \vspace{-0.5em}
\caption{Ablation study on the different architectures of video encoder. All models are pre-trained on HowTo100M dataset. 
} \vspace{-0.5em}
\label{tab:backbone_arch}
\end{table}

\cleardoublepage

\end{appendices}

\end{document}